\def\paperID{*****} 
\def\confName{CVPR}
\def\confYear{2026}

\author{First Author\\
Institution1\\
Institution1 address\\
{\tt\small firstauthor@i1.org}
\and
Second Author\\
Institution2\\
First line of institution2 address\\
{\tt\small secondauthor@i2.org}
}

\documentclass[10pt,twocolumn,letterpaper]{article}
\usepackage[T1]{fontenc}
\usepackage[utf8]{inputenc}
 \usepackage[pagenumbers]{cvpr} 
\usepackage{graphicx}
\usepackage{xcolor}
\usepackage{booktabs,multirow}
\usepackage[table]{xcolor}
\usepackage{threeparttable}
\usepackage{siunitx}
\newcommand{\acc}[2]{\num{#1 +- #2}}

\usepackage{float}
\usepackage{booktabs}
\usepackage{tabularx}
\usepackage{stfloats}
\usepackage{multirow}
\usepackage{comment}
\usepackage{verbatim}
\usepackage{xcolor}
\usepackage{amsthm}
\newtheorem{theorem}{Theorem}[section]  
\usepackage{graphicx}
\usepackage{subcaption}  
\definecolor{cvprblue}{rgb}{0.21,0.49,0.74}
\usepackage[pagebackref,breaklinks,colorlinks,allcolors=cvprblue]{hyperref}

\title{X-SAM: Boosting Sharpness-Aware Minimization with Dominant-Eigenvector Gradient Correction}

\author{
  Hongru Duan \quad Yongle Chen \quad Lei Guan\thanks{Corresponding author.} \\
  Taiyuan University of Technology \\
}

\begin{document}
\maketitle
\begin{abstract}
Sharpness-Aware Minimization (SAM) aims to improve generalization by minimizing a worst-case perturbed loss over a small neighborhood of model parameters. However, during training, its optimization behavior does not always align with theoretical expectations, since both sharp and flat regions may yield a small perturbed loss. In such cases, the gradient may still point toward sharp regions, failing to achieve the intended effect of SAM. To address this issue, we investigate SAM from a spectral and geometric perspective: specifically, we utilize the angle between the gradient and the leading eigenvector of the Hessian as a measure of sharpness. Our analysis illustrates that when this angle is less than or equal to ninety degrees, the effect of SAM’s sharpness regularization can be weakened.
Furthermore, we propose an explicit eigenvector-aligned SAM (X-SAM), which corrects the gradient via orthogonal decomposition along the top eigenvector, enabling more direct and efficient regularization of the Hessian’s maximum eigenvalue. We prove X-SAM’s convergence and superior generalization, with extensive experimental evaluations confirming both theoretical and practical advantages. 
\end{abstract}

\section{Introduction}\label{sec:intr}
Modern deep neural networks are often heavily overparameterized, and their loss landscapes tend to be highly non-convex. Consequently, identifying solutions with superior generalization ability among the many global minima has become a central challenge in machine learning. A substantial body of research has explored the mechanisms of model generalization from various perspectives \cite{wilson2017,foret2021,jastrzebski2018}. For example, in terms of the implicit properties of optimization algorithms, it has been shown that the inherent biases of stochastic gradient descent (SGD) and adaptive moment estimation (Adam) largely account for their differences in generalization performance \cite{wilson2017, soudry2018, kingma2015}; Other research has analyzed how the relationship between learning rate and batch size, as well as the use of large-batch training, influences both model structure and generalization ability~\cite{keskar2016}. Moreover, some explicit regularization methods leveraging second-order information have been proposed to enhance robustness in parameter space. For instance, Liu et al. (2022) introduce a regularization term based on a stochastic estimator of the Hessian trace, which penalizes sharp directions in the loss landscape and encourages the model to converge to flatter, more robust minima \cite{liu2022}. However, these approaches still struggle to directly explain the generalization behavior of models in complex loss landscapes, motivating researchers to revisit the problem from the perspective of loss geometry.

Against this backdrop, recent studies have shown that the geometric properties of the loss landscape in neural network parameter space—particularly the sharpness of local minima—are closely linked to model generalization performance \cite{keskar2016, jiang2019, xie2020}.
Specifically, neural networks converging to flatter minima tend to generalize better, whereas those reaching sharper minima are more prone to overfitting and poor robustness\cite{hochreiter1997}. 
This observation has motivated further investigation into how optimization algorithms implicitly explore and shape the geometry of the loss landscape during training, with the SAM method~\cite{foret2021, kaddour2022, yue2023} being particularly representative. SAM minimizes the worst-case loss within a local neighborhood of the parameter space, thereby explicitly controlling the sharpness of the solution to improve model performance.  

Intuitively, SAM does not merely attempt to minimize the loss at the current parameter point, but also aims to identify flat regions where the loss remains low under small perturbations, thereby finding flatter minima in the parameter space that are more robust and generalize better.
With this principle, SAM has demonstrated remarkable advantages across various domains. Whether in computer vision \cite{foret2021,li2024} or natural language processing tasks\cite{bahri2021}, SAM effectively alleviates overfitting and consistently outperforms existing state-of-the-art methods. However, recent studies show that the perturbed loss minimized by SAM can be low not only at flat minima but also at sharp minima \cite{zhuang2022, luo2024}, indicating that simply minimizing perturbed loss does not always guarantee a truly sharpness-aware solution. In other words, the SAM method may not always precisely control sharpness.

To investigate why minimizing the worst-case perturbed loss does not yield the desired sharpness-aware minimization effect, we conduct an empirical analysis of the gradient directions throughout the training process. We find that for a substantial portion of the training process, the angle between the gradient and the principal eigenvector of the Hessian remains nearly orthogonal (approximately $80^\circ$–$100^\circ$). In such cases, the gradient component along the principal direction is extremely small, making it difficult for the updates to effectively influence the dominant curvature, thereby hindering the reduction of the largest eigenvalue. When the angle lies within the range of $50^\circ$–$80^\circ$, the updates tend to move toward sharper regions of the loss landscape, further limiting the decrease of the dominant curvature. 

Based on this observation, and together with the PAC-Bayesian corollary~\cite{luo2024} that reveals the relationship between the largest eigenvalue of the Hessian and the model's generalization error, we propose a novel and efficient method, X-SAM. Our proposal leverages the dominant eigenvector of the Hessian as a reliable indicator of the sharpest curvature direction, and further refines the update rule by attenuating the gradient component projected onto this eigenvector. With this strategy, the largest eigenvalue of the Hessian is explicitly constrained, hence giving rise to finer control over sharpness. Experimental results demonstrate that X-SAM significantly enhances the generalization performance of the model compared to other four SAM-based approaches.

The main contributions of this paper can be summarized as follows:
\begin{itemize}
    \item Firstly, we analyze how the angle between the gradient and the Hessian's principal eigenvector in SAM affects the eigenvalues, and find that near-orthogonal directions fail to effectively reduce the maximum eigenvalue. Based on this observation, we propose a novel variant of SAM, termed X-SAM.
    \item Secondly, we theoretically demonstrate that X-SAM controls local curvature by removing the harmful gradient component along the principal eigenvector to avoid updates toward high-curvature regions, thereby effectively reducing the largest eigenvalue of the Hessian. Furthermore, we establish its theoretical convergence under non-convex stochastic optimization.
    \item Thirdly, we conducted extensive experiments to evaluate the effectiveness of X-SAM by comparing it with SAM~\cite{foret2021}, GSAM~\cite{zhuang2022}, WSAM~\cite{yue2023}, and EigenSAM~\cite{luo2024}. The results indicate that X-SAM consistently achieves superior or comparable generalization performance. For example, on the CIFAR-10 dataset with ResNet-18, X-SAM achieves a maximum accuracy gain of 2.08\% and an average improvement of 1.21\%, while on CIFAR-100, it achieves a maximum gain of 2.46\% and an average improvement of 0.84\%.

\end{itemize}

\section{Related Work}\label{sec:related}
SAM is an optimization algorithm that improves the generalization of deep neural networks by favoring parameters located in flat regions of the loss landscape~\cite{foret2021}. Since its introduction, SAM has garnered significant academic interest, prompting a substantial body of research dedicated to its theoretical foundations, practical implementations, and extensions~\cite{zhuang2022,yue2023,li2024,luo2024}.

\textbf{Extensions of SAM.}
Various extensions of SAM~\cite{foret2021} have been proposed to improve either its optimization performance or computational efficiency. GSAM~\cite{zhuang2022} employs a gradient projection approach to increase the loss at the original parameter location while keeping the perturbed loss unchanged, thereby minimizing the difference between the perturbed and unperturbed losses and implicitly emphasizing the weight of this loss difference.
WSAM~\cite{yue2023} introduces a weighted combination of the empirical loss and the sharpness measure, effectively integrating sharpness as a regularization term into the objective. 
Moreover, F-SAM~\cite{li2024} analyzes SAM’s perturbation and shows that the full-gradient component harms generalization. By removing it with Exponential Moving Average (EMA) estimation and keeping only mini-batch gradient noise, F-SAM improves both generalization and convergence in non-convex settings.
This allows a smooth trade-off between minimizing the empirical risk and seeking flatter minima. Recent theoretical studies have analyzed SAM and many of its variants~\cite{andriushchenko2022, si2024, oikonomou2025}.
Eigen-SAM~\cite{luo2024} provides a theoretical characterization of SAM dynamics via a third-order stochastic differential equation (SDE). 
By tracking the maximum eigenvalue of the Hessian and aligning the perturbation  with its principal eigenvector, Eigen-SAM can explicitly reduce the sharpest curvature during training, enhancing generalization.
Other recent studies explored the effect of sharpness regularization and simplicity bias in training deep networks~\cite{gatmiry2024a, gatmiry2024b}.

\textbf{Hessian Spectrum, Eigenvalues, and Generalization.} Recent studies have highlighted the importance of the Hessian spectrum—particularly its largest eigenvalue—in characterizing model generalization.
Lyu et al.~\cite{kailyu2022} interpret the dominant Hessian eigenvalue  as a direct indicator of sharpness, showing that smaller values often correspond to flatter minima and the dominant Hessian eigenvalue is sensitive to parameter rescaling. 
Building on this, Cohen et al.~\cite{cohen2021,sanarora2022} observed that during training, the dominant Hessian eigenvalue frequently oscillates around the stability threshold , implying that gradient descent dynamics naturally stabilize in regions of moderate curvature, which can affect generalization. Furthermore, Damian et al.~\cite{damian2022} show that higher-order dynamics generate a self-stabilization effect, which implicitly controls the dominant Hessian eigenvalue throughout training. 
Collectively, these studies suggest that the largest Hessian eigenvalue and its corresponding eigenvector are crucial for quantifying sharpness and understanding implicit regularization in deep learning.

\textbf{Theoretical Analyses and Convergence of SAM.}
A growing body of theoretical research has been devoted to understanding the optimization dynamics and convergence behavior of SAM~\cite{wilson2017,keskar2016, soudry2018}.
Foret et al.~\cite{foret2021} provide a PAC-Bayesian generalization bound, showing that the algorithm implicitly minimizes a local-entropy--based objective that favors flat minima.
Building on this foundation, Andriushchenko and Flammarion~\cite{andriushchenko2022} offer a deeper theoretical understanding by analyzing SAM under smooth non-convex settings.
They demonstrate that, under standard smoothness and bounded-variance assumptions, SAM achieves a convergence rate of $O(1/\sqrt{T})$. This result further indicates that SAM acts as an implicit regularizer, guiding the optimization process toward flatter regions of the loss surface. Later, Si and Yun~\cite{si2024} rigorously examine the convergence properties of practical SAM variants, showing that the standard implementation may not converge exactly to true stationary points due to the perturbation bias introduced by adversarial updates.
More recently, Oikonomou and Loizou~\cite{oikonomou2025} present a unified convergence analysis of SAM and its variants, deriving improved theoretical rates and clarifying how the interplay between step size, sharpness radius, and smoothness constants governs optimization stability.
In addition to convergence guarantees, the relationship between Hessian eigenvalues and generalization has attracted increasing attention.
Luo et al.~\cite{luo2024} propose Eigen-SAM, which interprets SAM’s training dynamics through a third-order stochastic differential equation (SDE) framework.
They theoretically demonstrated that aligning the perturbation direction with the principal Hessian eigenvector allows the optimizer to explicitly reduce the largest curvature during training, thereby improving generalization. This eigenvalue-based perspective bridges the gap between SAM’s empirical flatness-promoting behavior and its underlying geometric regularization mechanism.

\section{Preliminaries}

\subsection{Sharpness-Aware Minimization}

When optimizing neural networks using gradient descent, the traditional approach minimizes $f(w)$ along the direction $\nabla f(w)$, aiming to obtain a single-point parameter $w$ with low loss~\cite{drucker1991, song2024, deng2012, netzer2011}. However, as noted by Chaudhari ~\cite{chaudhari2019} al, standard training often converges to sharp regions of the loss surface, which impairs the model’s generalization ability. Different from the traditional method, Sharpness-Aware Minimization (SAM) constructs a neighborhood around the parameters to ensure performance stability under weight perturbations~\cite{ahn2023, barrett2020}.

Let $w \in \mathbb{R}^d$ denote the parameters of a deep neural network, and $f$ the non-convex empirical risk (loss function) over a dataset $D = \{(x_i, y_i)\}_{i=1}^N$. To seek solutions lying in flat minima, SAM enforces small loss within the neighborhood of $w$, leading to the following min–max optimization problem:
\begin{equation}
\min f_p(w) \triangleq \min \Big(\max_{\|\epsilon\|\le \rho} f(w+\epsilon)\Big),
\end{equation}
where $w+\epsilon$ represents the ``worst-case'' perturbed model within radius $\rho$.

Since the problem is highly non-convex, directly solving the inner maximization is challenging~\cite{barrett2020,compagnoni2023}. SAM adopts an approximation as 
\begin{equation}
\begin{aligned}
\epsilon_t 
= \arg \max_{\|\epsilon\|\le \rho} f(w_t+\epsilon) 
&\approx \arg \max_{\|\epsilon\|\le \rho} \Big( f(w_t) + \langle \nabla f(w_t), \epsilon \rangle \Big) \\
&\approx \arg \max_{\|\epsilon\|\le \rho} \langle g_t(w_t), \epsilon \rangle
\end{aligned}
\label{eq:eps_approx}
\end{equation}
where the first step is a first-order Taylor expansion, and the second step replaces the gradient $\nabla f(w_t)$ with the stochastic gradient $g_t(w_t)$ for mini-batch training~\cite{blanc2020, dagreou2024}.

Equation~\eqref{eq:eps_approx} has a closed-form solution:
\begin{equation}
\epsilon^{\text{SAM}}(w) = \rho \frac{\nabla f(w)}{\|\nabla f(w)\|}.
\end{equation}

Subsequently, SAM computes the gradient at the perturbed point $w_t+\epsilon_t$, and updates the parameters as
\begin{equation}
w_{t+1} = w_t - \eta \, g_t(w_t+\epsilon_t),
\end{equation}
where $\eta$ is the learning rate. In this way, SAM preserves gradient information while gradually converging to flat minima, thereby improving the generalization ability~\cite{ahn2023, barrett2023, song2024}.

\section{Methodology}
\subsection{ Empirical Analysis }

\begin{figure}[H]  
  \centering
  \includegraphics[width=1\linewidth]{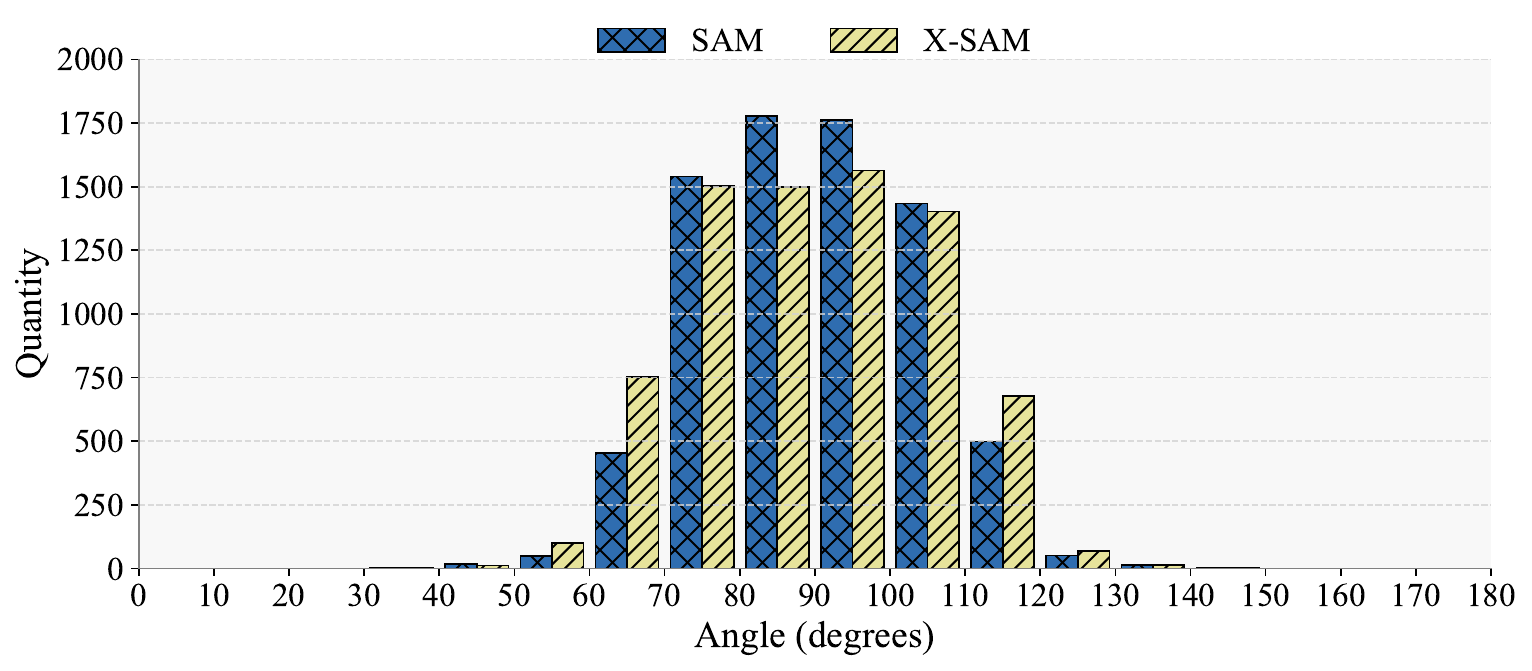}  
  \caption{Alignment statistics for a 6-layer CNN trained on CIFAR-10.
  The plots show the quantity distribution of the angle between the gradient (obtained from the second forward–backward pass of SAM) and the top Hessian eigenvector, which reflects the degree of alignment during training.}
  \label{fig:alpha}
\end{figure}

In this section, we perform an empirical study to analyze the actual update directions of model parameters during training. Specifically, we use the angle between the gradient and the principal eigenvector of the Hessian as an experimental metric to evaluate whether parameter updates follow the directions of maximum curvature on the loss surface. We trained a six-layer SimpleCNN model~\cite{jastrzebski2021} on the CIFAR-10 dataset and monitored this angle throughout the training process. The training was conducted with a batch size of 256, and for the purpose of analysis, we recorded the gradient-eigenvector angle every five batches to generate a statistical plot (as shown in Fig.~\ref{fig:alpha}), where the $x$-axis represents the angle between the gradient and the Hessian's principal eigenvector (in degrees), and the $y$-axis represents the frequency of each angle bin (count).
Concretely, after obtaining the gradient from the second backward pass and the Hessian's principal eigenvector $v$, we first flatten and L2-normalize the gradient tensor of each layer, and then concatenate them to form the full-model gradient vector. Similarly, the principal eigenvector of each layer is flattened and concatenated to form the full-model eigenvector. The angle between the gradient and the principal eigenvector is then computed by calculating the cosine similarity and converting it to degrees.

Based on the observation of Fig.~\ref{fig:alpha}, we see that for roughly 45\% of the training process, the angle remained between $80^\circ$ and $100^\circ$, indicating that the gradient is nearly orthogonal to the principal eigenvector.
This observation motivates a closer examination of the effect on the curvature along the principal eigenvector when the angle between the gradient and the eigenvector is either less than $80^\circ$ or nearly orthogonal.

To simplify the case where the angle is less than $80^\circ$, we assume that the gradient $g_t$ is aligned with the principal eigenvector of the Hessian: $g_t = \alpha v_1,\alpha > 0.$ The standard gradient descent update is then $\Delta w = -\eta g_t = -\eta \alpha v_1.$
The second-order change along the principal eigenvector direction is
\begin{equation}
\Delta w^\top H \Delta w = (\eta \alpha)^2 v_1^\top H v_1 = (\eta \alpha)^2 \lambda_1,
\end{equation}
indicating that the update along the direction of maximum curvature substantially increases the curvature along this direction, making the model more likely to enter or remain in sharp minima.

When the gradient is nearly orthogonal to the principal eigenvector $v_1$, i.e.,
$
\langle g_t, v_1 \rangle = \|g_t\| \cos \theta \approx 0,
$
the Hessian can be decomposed as
$
H = \sum_{i=1}^d \lambda_i v_i v_i^\top.
$
The contribution along the principal eigenvector \(v_1\) is
\begin{equation}
(\Delta w^\top H \Delta w)_{v_1} = \lambda_1 (\Delta w^\top v_1)^2=\lambda_1\eta^2 \langle g_t, v_1 \rangle^2.
\end{equation}

When $\langle g_t, v_1 \rangle \approx 0$, i.e., the gradient is nearly orthogonal to the principal eigenvector, the principal eigenvalue remains almost unchanged, indicating that reducing the dominant curvature becomes difficult, thereby limiting the effectiveness of flatness optimization.

\subsection{Gradient Decomposition}

Through the analysis of the aligned and orthogonal cases between the gradient and the principal eigenvector of the Hessian, we aim to enlarge the angle between them, thereby reducing updates along high-curvature directions and avoiding convergence to sharp minima. A natural idea is to directly modify the optimization objective as
\begin{equation}
\max_x \ \theta(w) \quad \text{or} \quad \min_w f(w) - \lambda \cos\theta(w).
\end{equation}
However, angle-based objectives are difficult to construct, because the derivative of $\theta(w)$ with respect to the parameters depends on both the gradient and the principal eigenvector, making it extremely challenging to compute in high-dimensional models. Moreover, there exists a potential gradient conflict: directly adding the gradient of $\cos\theta(w)$ to $\nabla f(w)$ may lead to conflicting update directions, where increasing the angle may unintentionally increase the original loss.

Therefore, we adopt the following gradient decomposition strategy
\begin{equation}
\nabla f(w) = \nabla f_\parallel + \nabla f_\perp,
\end{equation}
where $\nabla f_\parallel$ is the component along the principal eigenvector and $\nabla f_\perp$ is the orthogonal component. By decomposing the gradient, we can adjust the component along the principal eigenvector independently.



\subsection{The X-SAM Algorithm}

After computing the gradient component along the principal eigenvector, it can be used to correct the parameter update gradient in standard SAM, thereby directly constructing the proposed X-SAM. Specifically, at the $t$-th training iteration, the principal eigenvector is first intermittently estimated using the power iteration method to avoid the high computational cost of direct computation. The principal eigenvector is updated once every fixed number $p$ of mini-batch steps, performing $q$ iterations per estimation to ensure a reliable approximation.  

\begin{table}[H]
\label{alg:sam}
\centering
\setlength{\tabcolsep}{5pt}
\renewcommand{\arraystretch}{1.15}
\label{tab:sam}
\begin{tabularx}{0.45\textwidth}{X}  
\toprule
\textbf{Algorithm 1:} SAM Algorithm \\
\midrule
1. \textbf{for} $t = 1, 2, \ldots, T$ \textbf{do} \\
2. \quad Compute mini-batch loss $\nabla f(w)$ \\
3. \quad Compute SAM perturbation:\\ \quad\quad\quad\quad$\epsilon^{SAM}(w)=\rho \frac{\nabla f(w)}{\|\nabla f(w)\|}$ \\
4. \quad Generate adversarial point: $w^{adv} = w + \epsilon^{SAM}$ \\
5. \quad Compute gradient at perturbed point:\\\quad\quad\quad\quad $g^{adv} = \nabla f(w^{adv})$ \\
6. \quad Update parameters: $w_{t+1} = w_t - \eta  \nabla f(w^{adv})$ \\
7. \textbf{end for} \\
\bottomrule
\end{tabularx}
\end{table}

Next, the initial gradient is computed to obtain the perturbation direction, and the perturbed parameters are derived accordingly. To mitigate the influence of the gradient magnitude on the directional information, we normalize the perturbed gradient:
\begin{equation}
\hat{\nabla} f(w^{adv}) = \frac{\nabla f(w^{adv})}{\|\nabla f(w^{adv})\|}.
\end{equation}
Subsequently, this gradient is decomposed into two orthogonal components: one along the principal eigenvector direction and the other orthogonal to it:
\begin{equation}
\hat{\nabla} f(w^{adv})=\nabla f(w^{adv})_{\parallel}+\nabla f(w^{adv})_{\perp}.
\end{equation}

By subtracting the gradient component along the principal eigenvector, the angle between the gradient and the principal eigenvector is increased, causing the original gradient direction to deviate from the sharpest direction and making the update more biased toward flatter regions. 
Therefore, we correct the gradient by attenuating the component along the principal eigenvector, which suppresses the gradient along the sharpest directions of the loss surface while preserving the descent component along flatter directions; the model parameter \(w_t\) is then updated according to Eq.~(5) in Section~3.
\begin{equation}
\begin{split}
\nabla f(w)_{\text{X-SAM}} 
&= \hat{\nabla} f(w^{adv}) \\
&\quad - \alpha \cdot 
\operatorname{sign}(\langle \nabla f(w^{adv}), v \rangle) 
\nabla f(w^{adv})_{\parallel}
\end{split}
\end{equation}

The hyperparameter $\alpha$ controls the strength of this correction, and a sign factor $\text{sign}(\langle \nabla f(w^{adv}), \hat{v} \rangle)$ is incorporated to resolve directional ambiguity.
The overall procedure of X-SAM with SGD as the base optimizer is summarized in Algorithm 2.

\begin{table}[t]
\centering
\setlength{\tabcolsep}{5pt}
\renewcommand{\arraystretch}{1.15}
\begin{tabularx}{0.45\textwidth}{X} 
\toprule
\textbf{Algorithm 2:} X-SAM Algorithm \\
\midrule
        1. \textbf{for} $t_1 = 1, 2, \ldots, T$ \textbf{do} \\
        2. \quad\quad Compute mini-batch loss $f_{\gamma}(w_k)$ \\
        3. \quad\quad \textbf{if} $t_1 \bmod p = 1$: \\
        4. \quad\quad\quad Initialize $\hat{v}$ as a random unit vector \\
        5. \quad\quad\quad \textbf{for} $t_2 = 1, 2, \ldots, q$ \textbf{do} \\
        6. \quad\quad\quad\quad \textbf{Compute Hessian-vector product}: \\ 
           \quad\quad\quad\quad\quad\quad\quad\quad $\hat{v} = \nabla^2 f(w) \cdot \hat{v}$ \\
        7. \quad\quad\quad\quad \textbf{Normalize} $\hat{v}$: $\hat{v} \gets \hat{v}/\|\hat{v}\|$ \\
        8. \quad\quad\quad \textbf{return} $\hat{v}$ \\
        9. \quad\quad \textbf{end if}\\
        10.\quad\quad Compute the perturbation: \\
           \quad\quad\quad\quad $\epsilon^{SAM}(w)=\rho \frac{\nabla f(w)}{\|\nabla f(w)\|}$ \\
        11.\quad\quad Perturb the model: $w^{adv} = w_t + \epsilon^{SAM}$ \\
        12.\quad\quad Compute the unit gradient: \\
           \quad\quad\quad\quad\quad \textcolor{blue}{$\hat{\nabla} f(w^{adv}) = \dfrac{\nabla f(w^{adv})}{\|\nabla f(w^{adv})\|}$} \\
        13.\quad\quad Decompose the gradient along the eigenvector: \\
           \quad\quad\quad\quad\quad \textcolor{blue}{$\hat{\nabla} f(w^{adv})=\nabla f(w^{adv})_{\parallel}+\nabla f(w^{adv})_{\perp}$} \\
        14.\quad\quad Perform gradient correction: \\
           \quad\quad\quad \textcolor{blue}{$\nabla f(w_{\text{X-SAM}}) = \hat{\nabla} f(w^{adv}) - \alpha \cdot \operatorname{sign}(\langle \nabla f(w^{adv}), v \rangle) \nabla f(w^{adv})_{\parallel}$} \\
        15.\quad\quad Update parameters: $w_{t+1} = w_t - \eta \nabla f(w)_{\text{X-SAM}}$ \\
        16.\textbf{end for}\\
\bottomrule
\end{tabularx}
\label{alg:xsam}
\end{table}

From Algorithm 1 and Algorithm 2, it can be seen that X-SAM improves upon the standard SAM by introducing finer control over the gradient direction to enhance optimization. Specifically, standard SAM updates the parameters based on the worst-case perturbed loss, which can indirectly reduce the model's sensitivity to sharp regions. Unlike directly using the perturbed gradient (highlighted in blue in Algorithm~2), X-SAM first normalizes the gradient and explicitly decomposes it into components along the principal eigenvector and the orthogonal directions. By attenuating the gradient along the principal eigenvector, X-SAM increases the angle between the gradient and the sharpest curvature direction, biasing the updates toward flatter regions while preserving the descent component along flatter directions. This effectively suppresses updates along sharper directions, providing a more direct control over the largest eigenvalue of the Hessian and improving the model's generalization performance.

\begin{table*}[t]
\centering
\setlength{\tabcolsep}{5pt}
\renewcommand{\arraystretch}{1.15}
\caption{Summarization of Top-1 Accuracy (\%) results. The best accuracy results are indicated in bold.}
\label{tab:xsam-main}

\begin{threeparttable}
\begin{tabular}{
  l l
  S[table-format=2.2(2)]
  S[table-format=2.2(2)]
  S[table-format=2.2(2)]
  S[table-format=2.2(2)]
  >{\columncolor{gray!10}}S[table-format=2.2]
}
\toprule

\multirow[l]{2}{*}{\textbf{Datasets}} &\multirow[l]{2}{*}{\textbf{Methods}} &
\multicolumn{5}{c}{\textbf{Top-1 Accuracy (\%)}} \\

\cmidrule(lr){3-7}

\multicolumn{2}{c}{} &
\textbf{ResNet-18} &
\textbf{ResNet-50} &
\textbf{WRN-28-10} &
\textbf{AlexNet} &
\textbf{Avg.} \\
\midrule

\multirow[c]{5}{*}{\textbf{CIFAR-10}}
& SAM       & \acc{93.64}{0.30} & \acc{92.33}{0.54} & \acc{95.01}{0.18} & \acc{88.66}{0.56} & 92.41 \\
& WSAM      & \acc{93.01}{0.42} & \acc{91.68}{0.25} & \acc{94.23}{0.08} & \acc{90.08}{0.35} & 92.25 \\
& GSAM      & \acc{93.80}{0.24} & \acc{92.16}{0.39} & \acc{94.67}{0.42} & \acc{88.59}{0.13} & 92.31 \\
& Eigen-SAM & \acc{93.82}{0.17} & \acc{92.57}{0.29} & \acc{95.11}{0.10} & \acc{88.79}{0.19} & 92.57 \\
\rowcolor{gray!12}
& \textbf{X-SAM} & \bfseries\acc{94.71}{0.20} & \bfseries\acc{93.76}{0.07} & \bfseries\acc{95.53}{0.26} & \bfseries\acc{90.73}{0.09} & \bfseries 93.68 \\
\midrule
\multirow{5}{*}{\textbf{CIFAR-100}}
& SAM       & \acc{74.00}{0.24} & \acc{72.01}{1.67} & \acc{77.96}{0.15} & \acc{60.26}{0.35} & 71.06 \\
& WSAM      & \acc{73.65}{0.01} & \acc{71.04}{0.60} & \acc{77.19}{0.53} & \acc{65.90}{0.36} & 71.95 \\
& GSAM      & \acc{73.55}{0.09} & \acc{71.61}{0.54} & \bfseries\acc{78.15}{0.14} & \acc{60.14}{0.21} & 70.86 \\
& Eigen-SAM & \acc{73.74}{0.28} & \acc{71.90}{1.07} & \acc{78.20}{0.02} & \acc{60.49}{0.38} & 71.08 \\
\rowcolor{gray!12}
& \textbf{X-SAM} & \bfseries\acc{74.10}{0.37} & \bfseries\acc{74.08}{0.23} & \acc{78.23}{0.36} & \bfseries\acc{68.82}{0.01} & \bfseries 73.81 \\
\midrule
\multirow{5}{*}{\textbf{Fashion-MNIST}}
& SAM       & \acc{95.22}{0.14} & \acc{94.46}{0.44} & \acc{95.56}{0.10} & \acc{94.34}{0.03} & 94.90 \\
& WSAM      & \acc{95.32}{0.35} & \acc{94.57}{0.13} & \acc{95.40}{0.07} & \multicolumn{1}{c}{---} & 95.43 \\
& GSAM      & \acc{95.39}{0.03} & \acc{94.94}{0.20} & \bfseries\acc{95.68}{0.15} & \acc{94.61}{0.08} & 95.16 \\
& Eigen-SAM & \acc{95.36}{0.24} & \acc{94.76}{0.12} & \acc{95.61}{0.05} & \multicolumn{1}{c}{---} & \bfseries 95.24 \\
\rowcolor{gray!12}
& \textbf{X-SAM} & \bfseries\acc{95.52}{0.14} & \bfseries\acc{95.16}{0.04} & \acc{95.53}{0.04} & \bfseries\acc{94.75}{0.72} & \bfseries 95.24 \\

\bottomrule
\end{tabular}

\begin{tablenotes}[flushleft]\footnotesize
\footnotesize
\item \textbf{ “--” indicates that the model failed to converge.} 
\end{tablenotes}

\end{threeparttable}
\end{table*}

\section{Theoretical Analysis}\label{sec:the}
\subsection{Convergence Analysis}
\begin{theorem}
\label{thm:xsam-convergence}
Let $f:\mathbb{R}^d\to\mathbb{R}$ be $\beta$-smooth (i.e., $\nabla f$ is $\beta$-Lipschitz).
Fix $\alpha\in[0,2]$ and assume $\|v_t\|=1$ for all $t$, and define the projector
$
P_t:=I-\alpha v_tv_t^\top ,
\text{so that}
\|P_t\|_{\mathrm{op}}\le 1.
$
Let $\rho>0$ and define
$
g_t:=\nabla f(x_t),
h_t:=\nabla f(x_t+\rho\epsilon_t),
\hat h_t:=\nabla f(x_t+\rho\epsilon_t;\xi_t),
\tilde h_t:=P_t\hat h_t,
$
and update
$x_{t+1}=x_t-\eta\,\tilde h_t .$
Assume the stochastic gradient is conditionally unbiased with bounded conditional variance:
$\mathbb{E}[\hat h_t\mid x_t,\epsilon_t,v_t]=h_t,
\mathbb{E}\!\left[\|\hat h_t-h_t\|^2\mid x_t,\epsilon_t,v_t\right]\le \sigma^2 .$
Assume moreover that the perturbation satisfies $\|\epsilon_t\|^2\le 1+C_1\alpha^2$ almost surely for some $C_1>0$, and define
$D_\rho:=\sup_t \|g_t-h_t\|^2 \;\le\; \beta^2\rho^2(1+C_1\alpha^2).$
In particular, choosing
$\eta=\min\left\{\frac{1}{2\beta},\ \sqrt{\frac{\Delta}{\beta\sigma^2T}}\right\},$
there exist constants $c_1,c_2>0$ depending only on $c_0$ such that
\begin{equation}\label{eq:avg_grad_bound_rate}
\frac{1}{T}\sum_{t=0}^{T-1}\mathbb{E}\|\nabla f(x_t)\|^2
\;\le\;
c_1\left(\frac{\beta\Delta}{T}+\sqrt{\frac{\beta\Delta\sigma^2}{T}}\right)
+c_2(1+2\alpha^2)
\end{equation}
For non-convex stochastic optimization, Theorem D.2 shows that X-SAM achieves a convergence rate of $\mathcal{O}(1/\sqrt{T})$, 
which matches the convergence speed of standard SGD~\cite{kingma2015, soudry2018}. The detailed proof of Theorem 5.1 is shown in Appendix A. 
\end{theorem}

\subsection{Eigenvalue Decrease Enhances Generalization}
\begin{theorem}[decrease of the leading eigenvalue]\label{thm:lambda1_decrease}
Let $H$ be symmetric with eigengap
$\gamma:=\lambda_1(H)-\lambda_2(H)>0,$
and let $\Delta H:=\nabla^2 f(x+\Delta x)-\nabla^2 f(x)$. Define
$s:=\|\Delta H\|_{\mathrm{op}},$
and assume $s<\gamma$. Let $v_1$ be the unit leading eigenvector of $H$ and $v'$ be the unit leading
eigenvector of $H+\Delta H$.
Assume the uniform third-derivative bound
$\bigl|\nabla^3 f(z)[a,b,c]\bigr|\le M_3\|a\|\,\|b\|\,\|c\|,$
and $\Delta x=\eta\,\tilde h$. 
Moreover, the leading eigenvalue satisfies
\begin{equation}\label{eq:lambda1_upper}
\lambda_1(H+\Delta H)
\le
\lambda_1(H)+ v_1^\top \Delta H v_1
+
{\frac{4(\|H\|_{\mathrm{op}}+s)}{\gamma-s}}\,s.
\end{equation}
This result establishes a sufficient condition ensuring that the corrected update direction 
induces a local flattening effect on the loss surface, reducing the spectral sharpness.The detailed proof can be found in Appendix A.
\end{theorem}

\subsection{Orthogonality and Sharpness Reduction}
\begin{theorem}[first-order change of the leading eigenvalue]\label{thm:lambda1_change}
Let $f:\mathbb{R}^d\to\mathbb{R}$ be $\mathcal C^4$ in a neighborhood of $w_t$ and define
$H(w)=\nabla^2 f(w)$ and $H_t:=H(w_t)$.
Assume there exists $\rho>0$ such that the third and fourth derivatives are locally bounded:
$M_3:=\sup_{\|u-w_t\|\le \rho}\|\nabla^3 f(u)\|_{\mathrm{op}}<\infty,
M_4:=\sup_{\|u-w_t\|\le \rho}\|\nabla^4 f(u)\|_{\mathrm{op}}<\infty.$
Assume $H_t$ has a simple largest eigenvalue $\lambda_1(w_t)$ with unit eigenvector $v_1(w_t)$,
and let $\gamma:=\lambda_1(w_t)-\lambda_2(w_t)>0$.
Consider the update $\Delta w:=w_{t+1}-w_t=-\eta g_t$ with $\eta>0$ sufficiently small so that
$\|H(w_t+\Delta w)-H(w_t)\|_{\mathrm{op}}<\gamma/2.$
Assume further that
$\|(I-v_1v_1^\top)\nabla\lambda_1(w_t)\|\le \varepsilon,
\text{and}
v_1(w_t)^\top g_t=0.$
Then the leading eigenvalue increment $\Delta\lambda_1:=\lambda_1(w_{t+1})-\lambda_1(w_t)$ satisfies
\begin{equation}
|\Delta\lambda_1|
\le
\eta\varepsilon\|g_t\|
\;+\; O\!\bigl(\eta^2\|g_t\|^2\bigr)
\;+\; O\!\Bigl(\frac{\eta^2\|g_t\|^2}{\gamma}\Bigr)
\end{equation}
where the hidden constants depend only on $M_3,M_4$ and on the universal spectral constant in the perturbation remainder.
This result indicates that when the two directions are nearly orthogonal, the dominant first-order term vanishes, and under a small step size $\eta$ and a non-degenerate spectral gap $\gamma$, the SAM update has only a very limited effect on reducing the maximum eigenvalue. The detailed proof can be found in Appendix A.
\end{theorem}

\begin{figure*}[t]
  \centering
  \begin{minipage}[b]{1\linewidth}
    \centering
    \vspace{-0.1em} 
    \includegraphics[width=13cm]{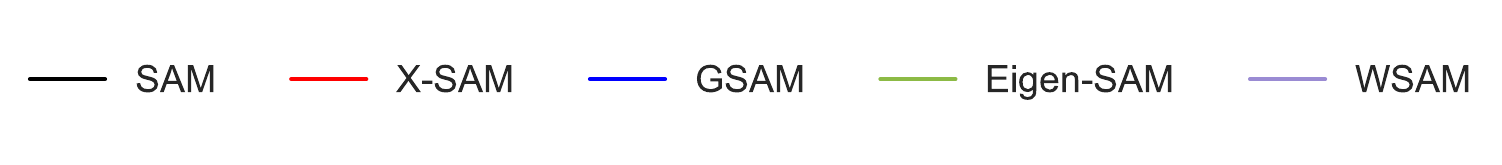} 
    \vspace{0.3em} 
  \end{minipage}

  \begin{subfigure}[b]{0.32\linewidth}
    \centering
    \includegraphics[width=\linewidth]{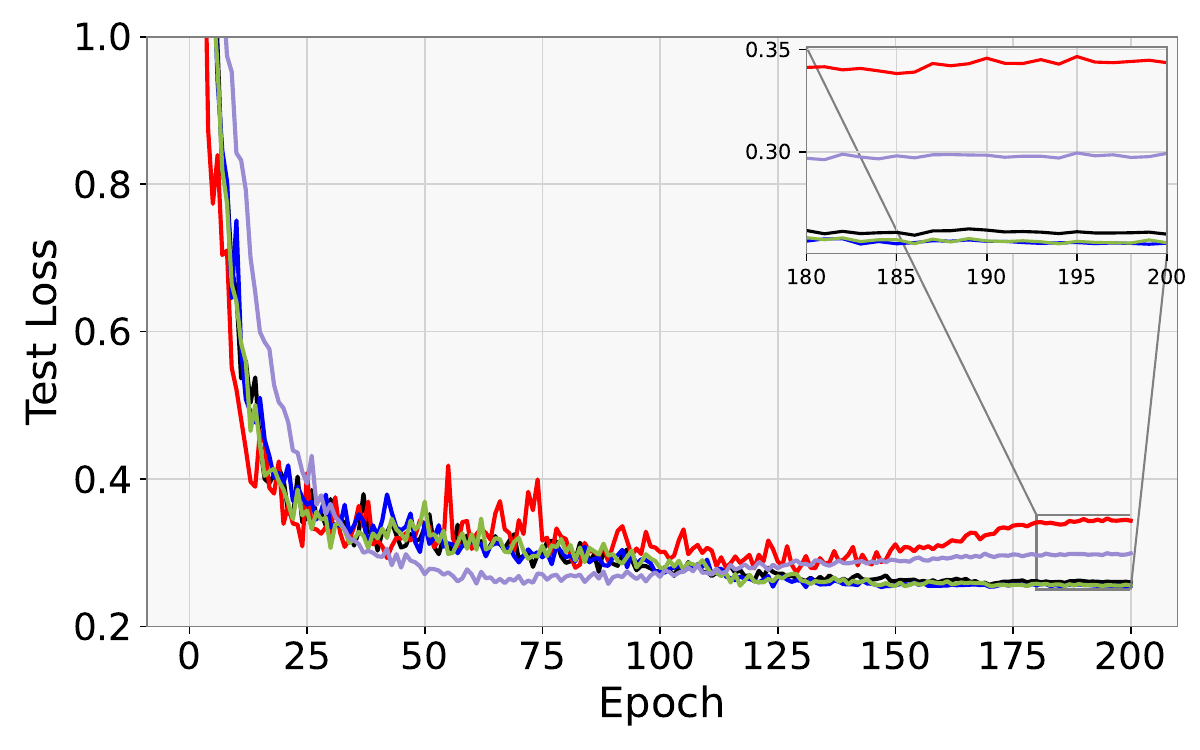}
    \caption{CIFAR-10}
    \label{fig:CIFAR-10-loss}
  \end{subfigure}
  \hspace{0.5em}
  \begin{subfigure}[b]{0.32\linewidth}
    \centering
    \includegraphics[width=\linewidth]{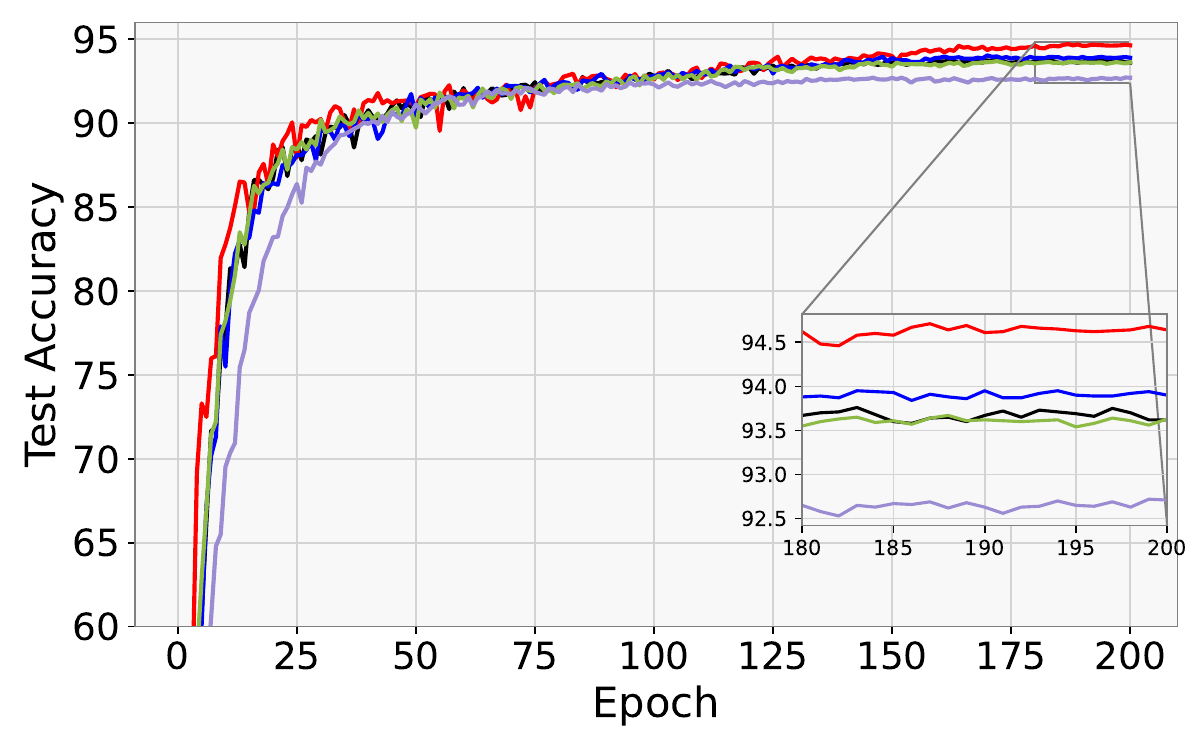}
    \caption{CIFAR-10}
    \label{fig:CIFAR-100-loss}
  \end{subfigure}
  \hspace{0.5em}
  \begin{subfigure}[b]{0.32\linewidth}
    \centering
    \includegraphics[width=\linewidth]{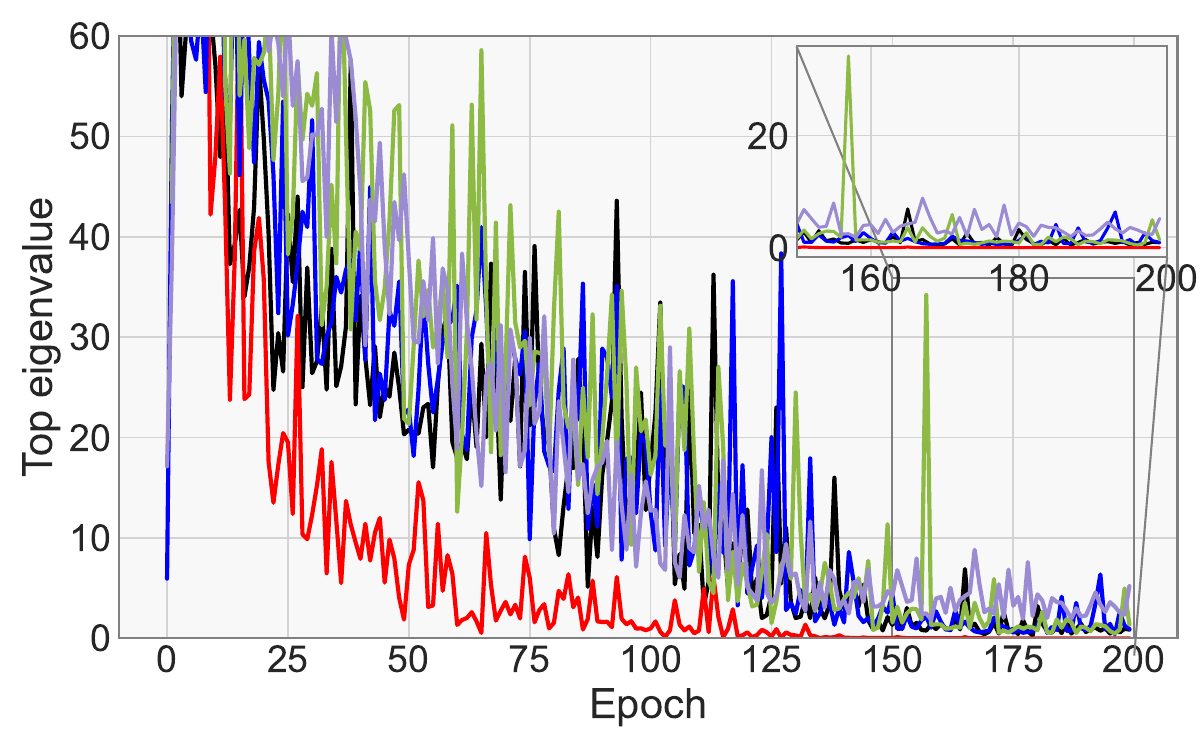} 
    \caption{Resnet-18}
    \label{fig:Fashion-MNIST-loss}
  \end{subfigure}

  \vspace{0.2em} 

  \begin{subfigure}[b]{0.32\linewidth}
    \centering
    \includegraphics[width=\linewidth]{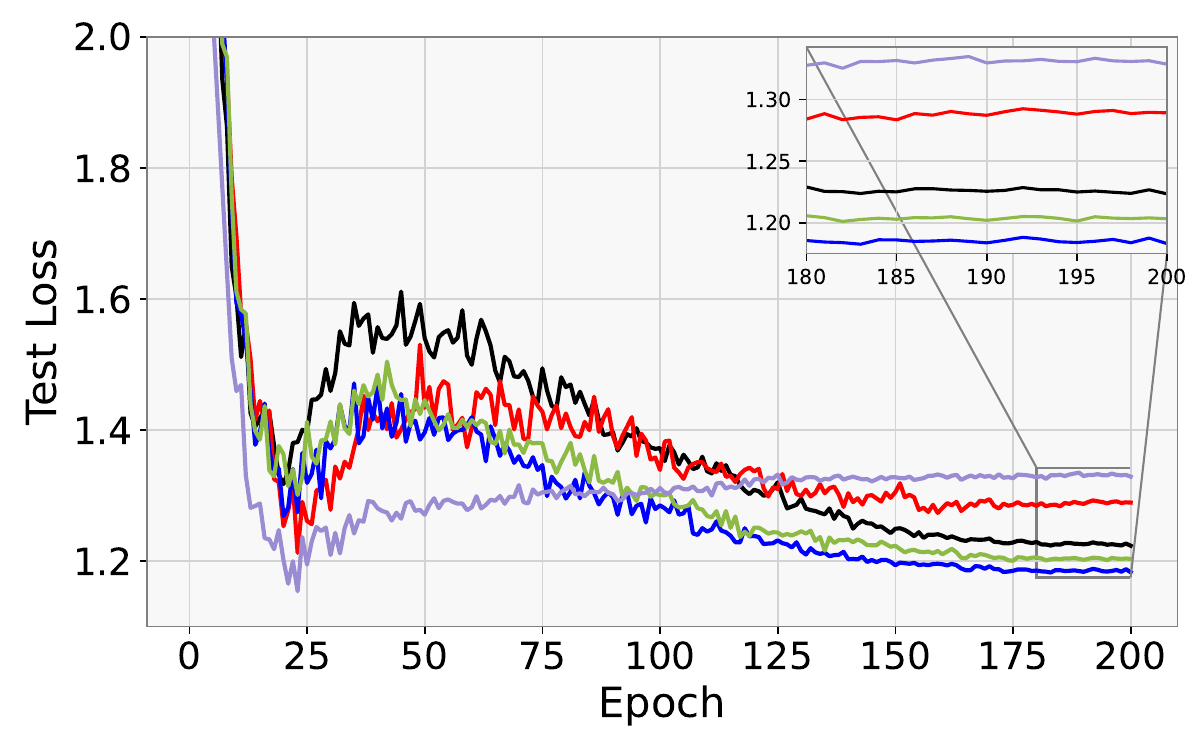}
    \caption{CIFAR-100}
    \label{fig:CIFAR-10-acc}
  \end{subfigure}
  \hspace{0.5em}
  \begin{subfigure}[b]{0.32\linewidth}
    \centering
    \includegraphics[width=\linewidth]{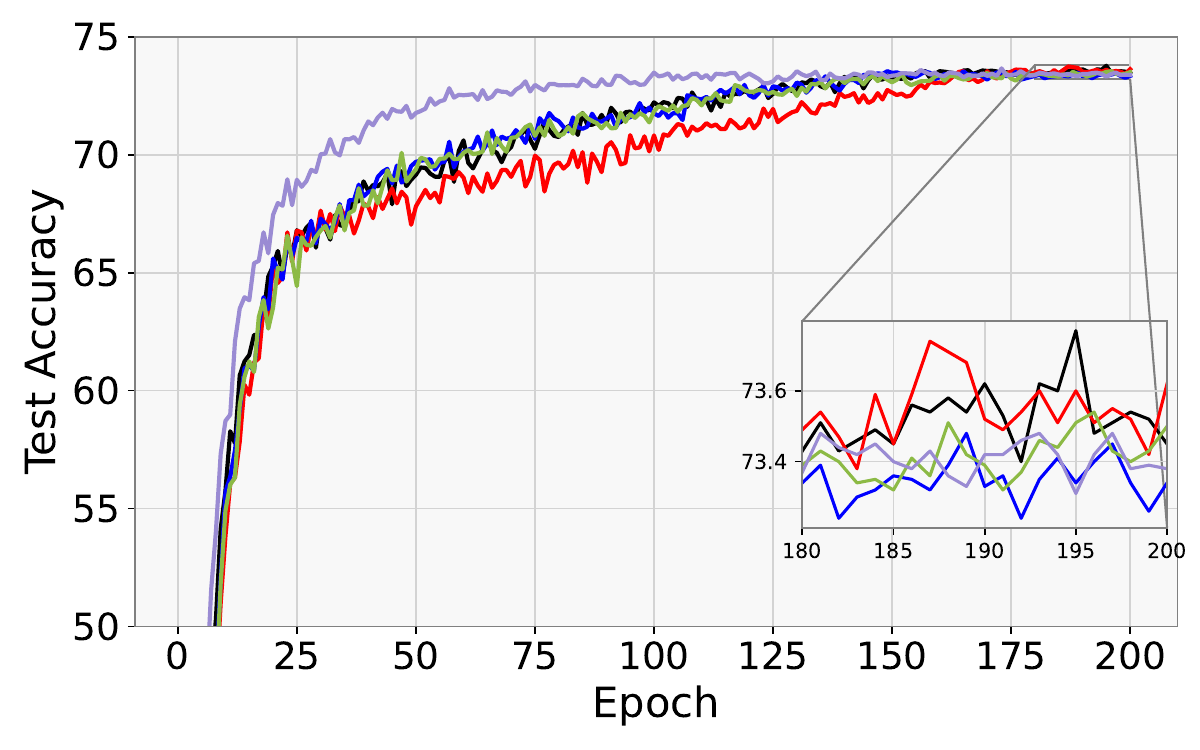}
    \caption{CIFAR-100}
    \label{fig:CIFAR-100-acc}
  \end{subfigure}
  \hspace{0.5em}
  \begin{subfigure}[b]{0.32\linewidth}
    \centering
    \includegraphics[width=\linewidth]{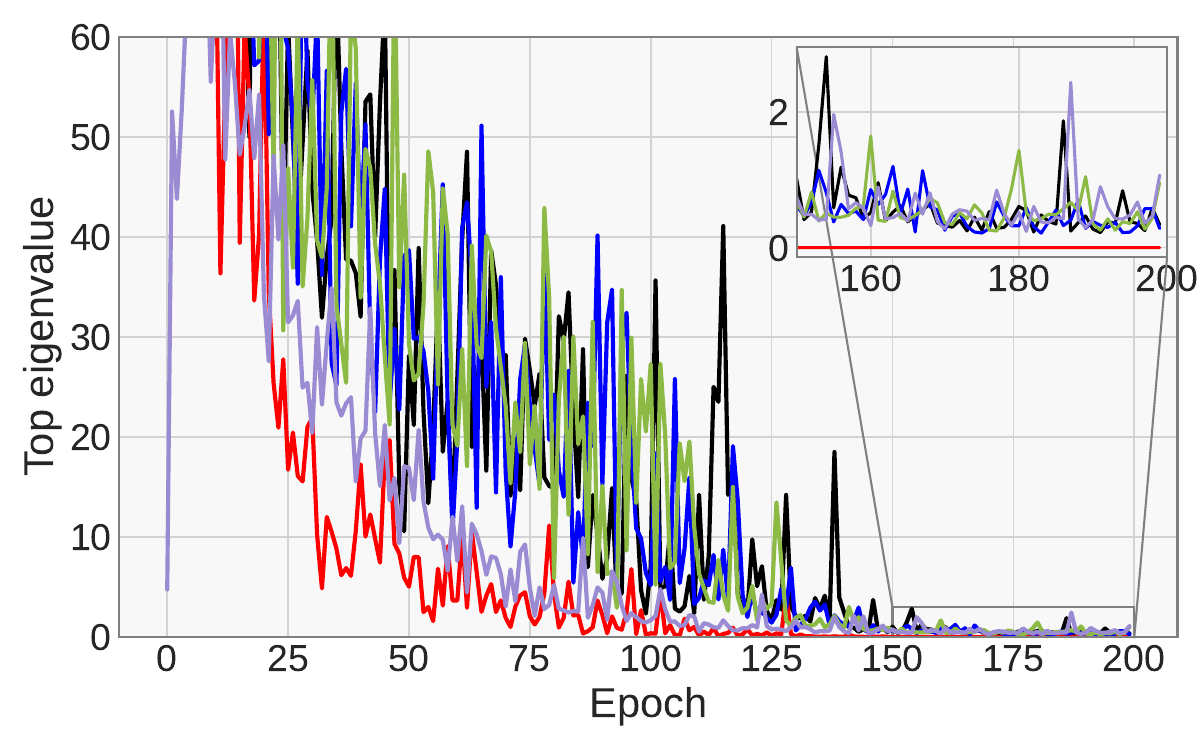} 
    \caption{WideResnet}
    \label{fig:Fashion-MNIST-acc}
  \end{subfigure}

  \caption{Experimental results when training ResNet-18 on CIFAR-10, CIFAR-100. Figs~\ref{fig:CIFAR-10-loss},~\ref{fig:CIFAR-10-acc}: Test loss vs. Epochs; Figs~\ref{fig:CIFAR-100-loss},~\ref{fig:CIFAR-100-acc}: Test accuracy vs. Epochs. Figs~\ref{fig:Fashion-MNIST-loss},~\ref{fig:Fashion-MNIST-acc}: Top eigenvalue vs. Epochs.
}
  \label{fig:train_test_compare}
\end{figure*}

\section{Experiments} \label{sec:experiment}
\subsection{Experimental Settings}

The experiments were conducted on a Linux-based server equipped with NVIDIA RTX 4090 GPUs. The software environment was configured with a Python 3.8 virtual environment and implemented using PyTorch 2.0.0 (CUDA 11.8). We performed experiments on the CIFAR-10, CIFAR-100~\cite{krizhevsky2009}, and Fashion-MNIST~\cite{xiao2017} datasets, using ResNet-18, ResNet-50~\cite{he2016}, AlexNet~\cite{krizhevsky2017}, and WideResNet-28-10~\cite{zagoruyko2016} as backbone models.  
We adopted SGD as the base optimizer, with an initial learning rate of 0.1 dynamically adjusted by a cosine schedule. The weight decay and momentum were set to $5 \times 10^{-5}$ and 0.9, respectively. The batch size was 256, and each model was trained for 200 epochs.  
For SAM and its variants, we used standard data augmentation techniques, including horizontal flipping, 4-pixel padding, and random cropping. The perturbation radii for SAM and Eigen-SAM~\cite{foret2021,luo2024} were set to 0.05 for CIFAR-10 and Fashion-MNIST, and 0.1 for CIFAR-100. For WSAM~\cite{yue2023}, the radii were set to 2 for CIFAR-10 and Fashion-MNIST, and 4 for CIFAR-100. Since GSAM~\cite{zhuang2022} was originally designed for fine-tuning, we adopted the same perturbation radii as SAM in our experiments. For X-SAM, we used the same perturbation radius as SAM and set $\alpha = 0.2$. The remaining parameter settings are provided in the Appendix A.  
All methods were trained under identical datasets, model architectures, data augmentation strategies, batch sizes, and training epochs. Each experiment was independently repeated three times to ensure the reliability and stability of the results.

\subsection{Performance Comparisons}
As shown in Table~\ref{tab:xsam-main}, X-SAM achieves the highest average test accuracy across all network architectures on all three datasets. For CIFAR-10, X-SAM shows an average improvement of approximately 1.3\%, with a maximum improvement exceeding 2.14\%, where the most significant gain is observed on AlexNet. For CIFAR-100, due to the outstanding performance of AlexNet, we consider it separately: excluding AlexNet, the average improvement is around 1.05\%, with a maximum improvement of up to 3.04\%; for the AlexNet architecture, the average improvement is also about 7.12\%, with a maximum improvement reaching 8.68\%. On Fashion-MNIST, X-SAM shows an average improvement of approximately 0.23\%, with a maximum improvement of up to 0.7\%. Overall, X-SAM demonstrates stable performance across different networks and datasets, with particularly notable improvements on shallow networks and more complex datasets.

Compared to SAM and its variants, X-SAM can more effectively control gradient updates along high-curvature directions, thereby improving model generalization. Specifically, in terms of test accuracy against several well-known SAM variants, X-SAM achieves an average improvement of approximately 0.8\% and a maximum improvement exceeding 2\% over GSAM; an average improvement of about 0.8\% and a maximum improvement exceeding 2\% over Eigen-SAM; and an average improvement of around 0.8\% with a maximum improvement exceeding 2\% over WSAM, demonstrating its robust advantage across multiple SAM variants.

Observing the training and test losses together with the largest Hessian eigenvalue (training losses are shown in Figures~\ref{fig:CIFAR-10-loss}, \ref{fig:CIFAR-10-acc}; test accuracies are shown in Figures~\ref{fig:CIFAR-100-loss}, \ref{fig:CIFAR-100-acc}; the largest eigenvalue is shown in Figure~\ref{fig:Fashion-MNIST-loss},\ref{fig:Fashion-MNIST-acc}; We find that X-SAM rapidly reduces the training loss within the first several epochs, while its largest Hessian eigenvalue decreases faster and exhibits smaller oscillations than other competing methods. This phenomenon suggests that X-SAM can more effectively restrict updates along high-curvature directions, enabling the optimization process to leave sharp regions earlier and enter a flatter basin, thereby improving training stability.
In the later stage of training, although the training loss of X-SAM slightly increases, the largest eigenvalue consistently remains at a low level, indicating that X-SAM does not merely pursue empirical risk minimization. Instead, it implicitly regularizes the geometric properties of the solution during optimization: by maintaining low curvature, it enhances the model's robustness to parameter perturbations.

\subsection{Hessian Spectrum Investigation}
\begin{figure}[H]
  \centering
  \includegraphics[width=0.6\linewidth]{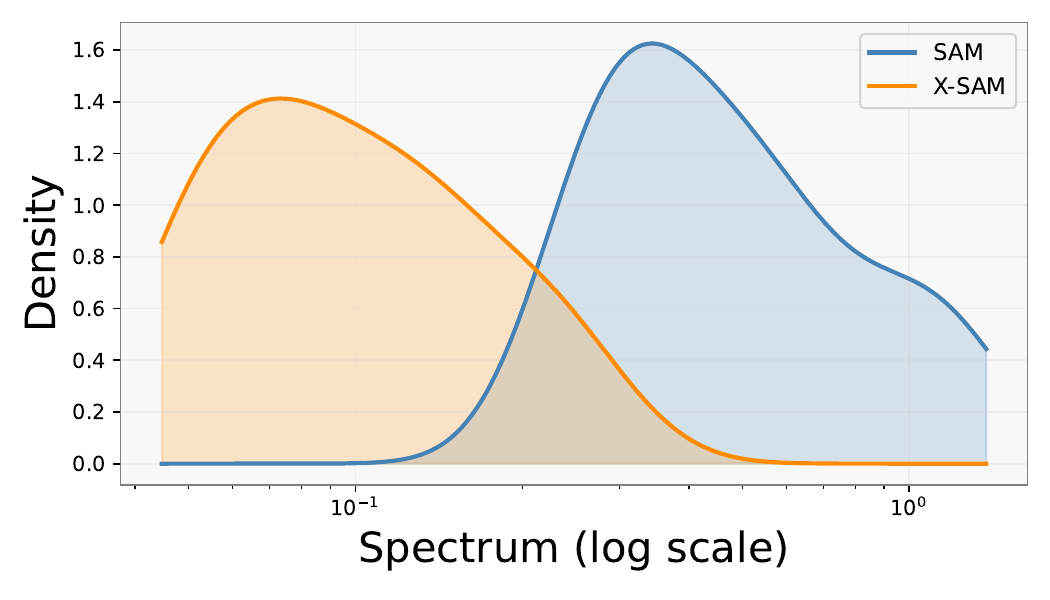} 
  \caption{Hessian spectrum at the end of training ResNet-18 on CIFAR-100.}
  \label{fig:spectrum}
\end{figure}

We analyze the impact of training with SAM and X-SAM on the Hessian spectrum of the model. The experiment is conducted on a ResNet-18 model trained on the CIFAR-100 dataset, where the top $k$ eigenvalues of the Hessian matrix $\nabla^2 f(w)$ are computed at the second epoch and visualized. The results, as shown in Figure~\ref{fig:spectrum}, show that compared with SAM, X-SAM consistently produces a smaller largest eigenvalue and a smaller trace, with a higher concentration of eigenvalues near zero. This indicates that X-SAM guides the optimization towards flatter regions of the loss surface, confirming the effectiveness of the proposed gradient correction mechanism and explaining why Eigen-SAM achieves better generalization performance than SAM.

\subsection{Hyperparameter Sensitivity Investigation of \texorpdfstring{$\alpha$}{alpha}}
\begin{figure}[H]
  \centering
  \includegraphics[width=0.6\linewidth]{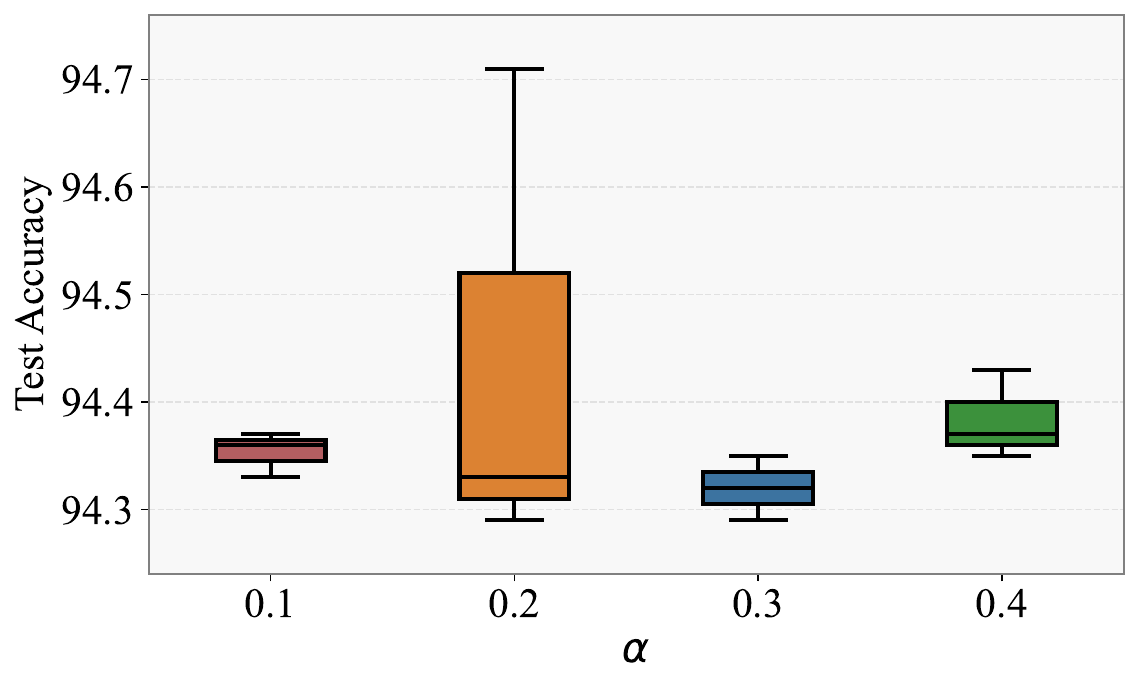}  
  \caption{Test accuracy of ResNet-18 when training on CIFAR-10 under different values of $\alpha$ for X-SAM.}
  \label{fig:alpha_sensitivity}
\end{figure}

We explore how the hyperparameter $\alpha$ influences the performance of X-SAM by running experiments on the CIFAR-100 dataset with the ResNet-18 architecture. Different values of $\alpha$ are evaluated, as illustrated in Figure~\ref{fig:alpha_sensitivity}. The results indicate that when $\alpha = 0.2$, the model achieves the highest test accuracy among all tested settings, showing a 1.07\% improvement over standard SAM. This suggests that selecting an appropriate $\alpha$ can significantly enhance performance. Moreover, the test accuracy remains relatively stable when $\alpha$ varies within the range $[0.1, 0.4]$, indicating that $\alpha$ is a fairly robust hyperparameter.

\section{Conclusion}
In this work, we conduct an in-depth study of gradient directions in SAM optimization. By analyzing the angle between the gradient and the dominant eigenvector of the Hessian, we find that when this angle is less than ninety degrees, the SAM optimization tends to converge to sharp regions. Based on this observation, we propose X-SAM which intermittently estimates the dominant eigenvector of the Hessian, decomposes the gradient along this direction, and suppresses its parallel component, thereby enlarging the angle between the gradient and the eigenvector while explicitly constraining the largest eigenvalue. It should be noted that the computational cost of eigenvector estimation remains non-negligible in the current implementation. In future work, we aim to reduce the computational overhead of X-SAM and extend its application to training and fine-tuning of large-scale models

\clearpage
{
    \small
    \bibliographystyle{ieeenat_fullname}
    \bibliography{main}
}               

\end{document}